\documentclass[runningheads]{llncs}
\usepackage[T1]{fontenc}
\usepackage{graphicx,verbatim}

\usepackage{xspace}
\usepackage{amsmath}
\usepackage{amssymb}

\usepackage[ruled,linesnumbered]{algorithm2e}
\usepackage[colorlinks,linkcolor=blue]{hyperref}
\usepackage[table]{xcolor}
\usepackage{multirow}
\usepackage{mathrsfs}
\usepackage{booktabs}
\usepackage{threeparttable}
\usepackage{bbm}
\usepackage{marvosym}
\usepackage{caption}
\begin{document}
\title{Brain-Adapter: A Dual-Stream Vision-Language MIL Framework for Comprehensive 3D CT Diagnosis of Acute Intracranial Pathologies}
\titlerunning{Brain-Adapter}
%
\author{Zhenyu Yi\inst{1} \and
Zhiyun Song\inst{2} \and
Yusong Sun\inst{1} \and
Zelin Liu\inst{1} \and
Manman Fei\inst{1} \and
Zhenhao Li\inst{1} \and
Jiaxuan Zhao\inst{1} \and
Xu Han\inst{1} \and
Lichi Zhang\textsuperscript{\textrm{(\Letter)}}\inst{1}}
%
%
\authorrunning{Z. Yi et al.}
%
\institute{School of Biomedical Engineering, Shanghai Jiao Tong University, Shanghai, China\\
\email{lichizhang@sjtu.edu.cn} \and
Department of Computing, Imperial College London, London, UK}
  
\maketitle              
\begin{abstract}
Automated diagnosis of 3D brain CT scans is essential for critical care, yet it remains challenging due to the heavy reliance on manual annotations and the limited semantic understanding of conventional models. While 2D foundation vision-language models (VLMs) have shown remarkable generalization, effectively transferring their representational power to 3D volumes remains an open problem. In this paper, we propose Brain-Adapter, a novel dual-stream multiple instance learning (MIL) framework that leverages pre-trained 2D biomedical VLMs and raw diagnostic reports for robust scan-level multi-label classification. Specifically, we introduce a Text-Conditioned Attention (TCA) mechanism, utilizing raw diagnostic sentences as semantic queries to dynamically align visual cues with specific disease concepts. Concurrently, a parallel visual MIL stream captures global scan characteristics, supervised by structured labels extracted via a Large Language Model (LLM). To ensure representation coherence, a consistency constraint enforces synergy between the two streams. During inference, an Uncertainty-Aware Refinement (UAR) module dynamically calibrates and fuses these dual-stream predictions to resolve ambiguous cases. Extensive experiments demonstrate that our method significantly outperforms state-of-the-art 3D models and standard MIL approaches. By eliminating the reliance on dense annotations, Brain-Adapter provides a highly scalable and clinically viable solution for 3D acute intracranial pathology analysis.

\end{abstract}

\section{Introduction}
Acute intracranial pathologies encompass a highly heterogeneous spectrum of lesions, presenting as critical medical emergencies that necessitate immediate triage~\cite{bg1,bg2}. Non-contrast computed tomography (NCCT) serves as the frontline imaging modality and is routinely reconstructed into thick-slice volumes for rapid evaluation~\cite{bg3}. However, the complex and diverse nature of brain lesions, coupled with the prohibitive cost of expert annotations, creates a formidable bottleneck for clinical scalability, motivating annotation-efficient and weakly supervised paradigms across medical AI~\cite{hu2024sali,hu2025monobox,zhao2024ultrasound,hu2026samix}. Furthermore, traditional unimodal deep learning models are fundamentally constrained in this domain; they not only demand massive, exhaustively annotated datasets but also lack the broad semantic understanding required to interpret open-ended diagnostic scenarios~\cite{cq500,lopez2022deep,neethi2024comprehensive}.

Medical foundation vision-language models (VLMs)~\cite{medclip,pubmedclip,biomedclip} offer a promising paradigm to circumvent manual labeling, exhibiting remarkable semantic understanding and zero-shot transferability~\cite{huang2021gloria,bommasani2021opportunities,xiang2025vision}. Yet, a fundamental dimensionality gap hinders their direct application: native 3D VLMs remain rare due to the severe scarcity of paired 3D-text corpora. Multiple Instance Learning (MIL) naturally bridges this gap by treating thick-slice 3D volumes as "bags" of 2D slices~\cite{dinoatten3d,berrimi2024multi,lopez2022deep}, enabling the transfer of 2D VLM representations to 3D tasks~\cite{liu2025revisiting}. Nevertheless, standard visual MIL approaches operate entirely within the visual domain without explicit semantic guidance. Consequently, they often struggle to pinpoint subtle pathologies, effectively attempting to localize lesions without prior knowledge of \textit{what} specific diagnostic concepts to search for~\cite{mil1,mil2,mil3}.

To address these limitations, we propose Brain-Adapter, a unified dual-stream MIL framework that adapts 2D VLMs for 3D acute intracranial pathology diagnosis, optimized entirely using raw diagnostic reports as supervision. We first leverage a Large Language Model (LLM) to automatically distill free-text reports into structured logic-sets. An \textbf{Open-set Alignment} branch utilizes a Text-Conditioned Attention (TCA) mechanism to retrieve pathology-relevant visual features via contrastive learning. And a \textbf{Logic-set Supervision} branch acts as a visual self-learning stream, employing an attention-based MIL~\cite{abmil} head supervised by the LLM-extracted labels to autonomously capture intrinsic morphological patterns across slices. A consistency constraint harmonizes these streams to ensure representation coherence, reconciling explicit semantic guidance with implicit visual evidence. During inference, an \textbf{Uncertainty-Aware Refinement (UAR)} strategy dynamically fuses the dual-stream predictions, calibrating visual outputs with text-guided evidence in ambiguous cases. 
We comprehensively evaluate Brain-Adapter on a real-world clinical dataset comprising complex cases. Experiments demonstrate that our method significantly outperforms state-of-the-art 3D models and standard 2D MIL approaches in multi-label classification. Moreover, to evaluate cross-domain transferability, a zero-shot abnormality detection task was conducted on the external CQ500~\cite{cq500} dataset, further confirming the robust generalization capabilities of our framework.

\section{Method}
\begin{figure}[t]
\centering
\includegraphics[width=0.9\textwidth]{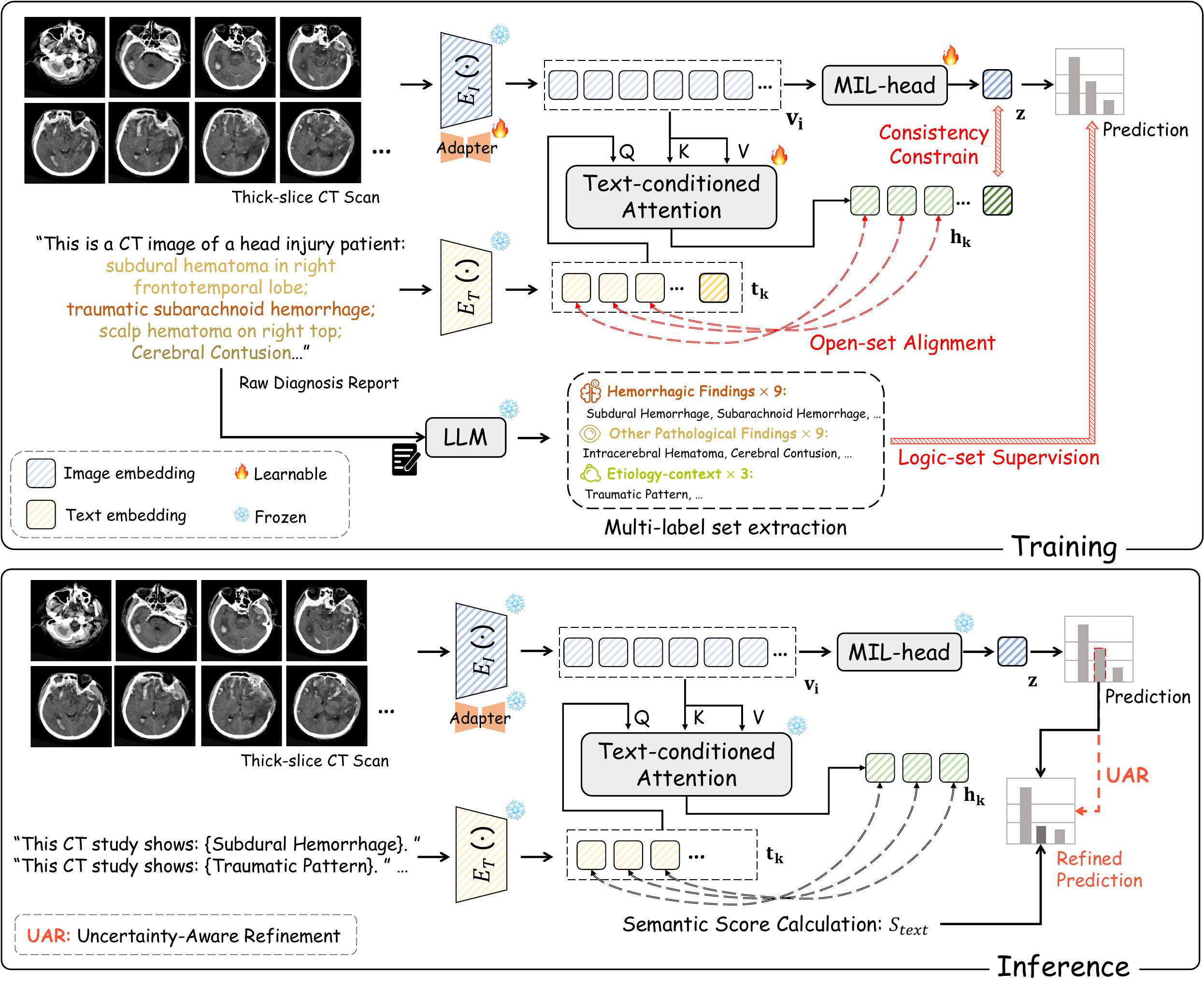}
\caption{Overview of Brain-Adapter. \textbf{Training (Top):} LoRA-adapted slice features feed two pathways. TCA matches text embeddings with relevant slices for \textit{Open-set Alignment}, while ABMIL aggregates features for \textit{Logic-set Supervision} using LLM-extracted labels. A consistency loss synchronizes global representations $\mathbf{z}$ and $\mathbf{h}_0$. \textbf{Inference (Bottom):} UAR dynamically fuses MIL predictions with prompt-based semantic scores ($S_{text}$) to rectify ambiguous cases.}
\label{fig1}
\end{figure}

We propose \textbf{Brain-Adapter}, a framework designed to adapt the frozen 2D VLM for 3D volumetric analysis. As illustrated in Fig.~\ref{fig1}, the architecture bifurcates into two synergistic branches during the training phase: a TCA pathway for fine-grained Open-set Alignment, and an Attention-based MIL pathway guided by Logic-set Supervision. And during the inference phase, an UAR mechanism dynamically fuses the outputs of both branches to enhance diagnostic reliability.

\subsection{Preliminaries: Structured Logic Extraction and Feature Encoding}
Given a thick-slice CT scan $X = \{x_i\}_{i=1}^N$ of a patient and a raw diagnostic report $R$, the initial objective is to obtain structured supervision and robust visual embeddings.

\noindent\textbf{LLM-driven logic-set extraction.}
Raw diagnostic reports $R$ consist of the physician's final "impression", typically presenting as unstandardized, semicolon-separated diagnostic terms. We leverage an LLM to parse this terminological diversity into a structured multi-label set $\mathcal{Y} \in \{0, 1\}^C$ ($C=21$). Encompassing Hemorrhagic Findings, Other Pathologies, and Etiology Context, this \textit{Logic-set} provides the explicit ground truth for our supervised branch.

\noindent\textbf{Feature encoding.}
The contrastive pretraining stage of vision-language models enables the alignment of visual and textual latent spaces. To transfer the 2D biomedical representation capabilities to 3D CT volumes while preserving robust semantic reasoning, Low-Rank Adaptation (LoRA)~\cite{hu2022lora} is applied to the self-attention blocks within the vision encoder, whereas the text encoder remains strictly frozen. The sequence of slice features is processed by the vision encoder $E_V$ to generate visual embeddings $V = [\mathbf{v}_1, \dots, \mathbf{v}_N] \in \mathbb{R}^{N \times d}$, where $\mathbf{v}_i = E_V(x_i)$. For the textual modality, both the complete raw diagnostic report and the subdivided fine-grained pathological descriptions are processed by the text encoder $E_T$. This yields global and fine-grained textual features $T = [\mathbf{t}_0, \mathbf{t}_1, \dots, \mathbf{t}_K] \in \mathbb{R}^{(K+1) \times d}$, where $\mathbf{t}_0$ represents the embedding of the global report and $\{\mathbf{t}_k\}_{k=1}^K$ denotes the embeddings of the fine-grained sentences.

\subsection{Open-set Alignment via Text-Conditioned Attention}
Inspired by the clinical practice where radiologists examine specific slices to identify distinct pathologies, a \textbf{Text-Conditioned Attention (TCA)} mechanism is proposed to explicitly align visual regions with fine-grained pathological descriptions. For a specific pathological description $k \in \{1, \dots, K\}$, the corresponding text embedding $\mathbf{t}_k$ serves as the \textit{Query}, while the slice features $V$ act as the \textit{Keys} and \textit{Values}. The pathology-specific visual representation $\mathbf{h}_k$ is aggregated as follows:
\begin{equation}
    A_k = \text{Softmax}\left(\frac{\mathbf{t}_k V^\top}{\sqrt{d}}\right), \quad \mathbf{h}_k = A_k V
\end{equation}
where $A_k \in \mathbb{R}^{1 \times N}$ represents the attention map indicating the relevance of each slice to pathology $k$.

To ensure that $\mathbf{h}_k$ accurately captures the semantics of pathology $k$, an \textbf{InfoNCE} loss~\cite{infonce} is employed for fine-grained alignment. The objective maximizes the similarity between the aggregated visual feature $\mathbf{h}_k$ and the corresponding text embedding $\mathbf{t}_k$, while distancing it from the text embeddings of other pathologies $\mathbf{t}_{j}$ ($j \neq k$) within the set:
\begin{equation}
    \mathcal{L}_{align} = - \sum_{k=1}^K \log \frac{\exp(\text{cos}(\mathbf{h}_k, \mathbf{t}_k) / \tau)}{\sum_{j=1}^K \exp(\text{cos}(\mathbf{h}_k, \mathbf{t}_j) / \tau)}
\end{equation}
where $\text{cos}(\cdot)$ denotes cosine similarity and $\tau$ is a temperature parameter. This formulation forces the model to attend to slices that strictly correspond to the provided textual description.

\subsection{Logic-set Supervision for Global Aggregation}
To obtain a holistic scan-level representation, we employ a gated Attention-based MIL (ABMIL)~\cite{abmil} aggregator. Capturing complex non-linear relations, the attention weight $a_i$ for slice $i$ is formulated as:
\begin{equation}
    a_i = \frac{\exp\{\mathbf{w}^\top (\tanh(\mathbf{W} \mathbf{v}_i) \odot \sigma(\mathbf{U} \mathbf{v}_i))\}}{\sum_{j=1}^N \exp\{\mathbf{w}^\top (\tanh(\mathbf{W} \mathbf{v}_j) \odot \sigma(\mathbf{U} \mathbf{v}_j))\}}
\end{equation}
where $\mathbf{W}, \mathbf{U} \in \mathbb{R}^{L \times d}$ and $\mathbf{w} \in \mathbb{R}^L$ are learnable parameters, $\odot$ is element-wise multiplication, and $\sigma(\cdot)$ is the sigmoid function. The global patient representation is then aggregated as $\mathbf{z} = \sum_{i=1}^N a_i \mathbf{v}_i$.

Inspired by SuperCLIP~\cite{zhao2025superclip}, we explicitly supervise $\mathbf{z}$ using the structured logic-set $\mathcal{Y}$. A linear head maps $\mathbf{z}$ to class logits $\mathcal{P} = (p_1, \dots, p_C)$. To mitigate the severe class imbalance inherent in clinical data, we optimize this multi-label classification via Asymmetric Loss (ASL)~\cite{asl} instead of standard BCE. Given the predicted probability $P_c = \sigma(p_c)$, the logic-set supervision loss is:
\begin{equation}
    \mathcal{L}_{logic} = \frac{1}{C} \sum_{c=1}^C \left[ y_c \, \ell_1(P_c) + (1 - y_c) \, \ell_0(P_c) \right]
\end{equation}
where $\ell_1(P_c) = -(1 - P_c)^{\lambda_1} \log(P_c)$ and $\ell_0(P_c) = -P_c^{\lambda_0} \log(1 - P_c)$. Here, $y_c \in \{0, 1\}$ is the ground truth from $\mathcal{Y}$, and $\lambda_1, \lambda_0$ are the focusing parameters.

Finally, to synchronize the visual and text-guided pathways, a consistency constraint aligns $\mathbf{z}$ with the global report-conditioned representation $\mathbf{h}_0$ (derived by feeding the global report embedding $\mathbf{t}_0$ into the TCA module):
\begin{equation}
    \mathcal{L}_{cons} = 1 - \frac{\mathbf{z}^\top \mathbf{h}_0}{\|\mathbf{z}\| \|\mathbf{h}_0\|}
\end{equation}

\subsection{Uncertainty-Aware Refinement (UAR)}
To rectify ambiguous MIL predictions during inference, we propose UAR to dynamically leverage the fine-grained semantic grounding of the TCA branch.

For a specific class $c$, let $P_c = \sigma(p_c) \in [0, 1]$ denote the predicted probability from the MIL branch. We define an uncertainty score $u_c$:
\begin{equation}
    u_c = 1 - |2 P_c - 1|^\alpha
\end{equation}
where $\alpha \ge 1$ controls sensitivity, peaking ($u_c \to 1$) when the prediction is highly ambiguous ($P_c \approx 0.5$). 

To compute the complementary semantic score, we insert the disease label into the prompt \textit{"This CT study shows: \{disease\}."} to extract the text embedding $\mathbf{t}_c$. Using $\mathbf{t}_c$ as a query in the TCA module, we retrieve the pathology-specific visual representation $\mathbf{h}_c$. The probability-aligned semantic score is then computed as $S_{\text{text}}^{(c)} = \sigma( \text{cos}(\mathbf{h}_c, \mathbf{t}_c) / \tau )$. Finally, the refined prediction dynamically fuses both branches via an adaptive convex combination:
\begin{equation}
    P_{\text{refined}}^{(c)} = (1 - \lambda \cdot u_c) P_c + (\lambda \cdot u_c) S_{\text{text}}^{(c)}
\end{equation}
where $\lambda \in [0, 1]$ is a scaling factor. This dynamic formulation ensures that the text-alignment branch is exclusively engaged to rectify predictions when the visual discriminative logic remains inconclusive.

\section{Experiments and Results}

\begin{figure}[t]
\centering
\includegraphics[width=0.7\textwidth]{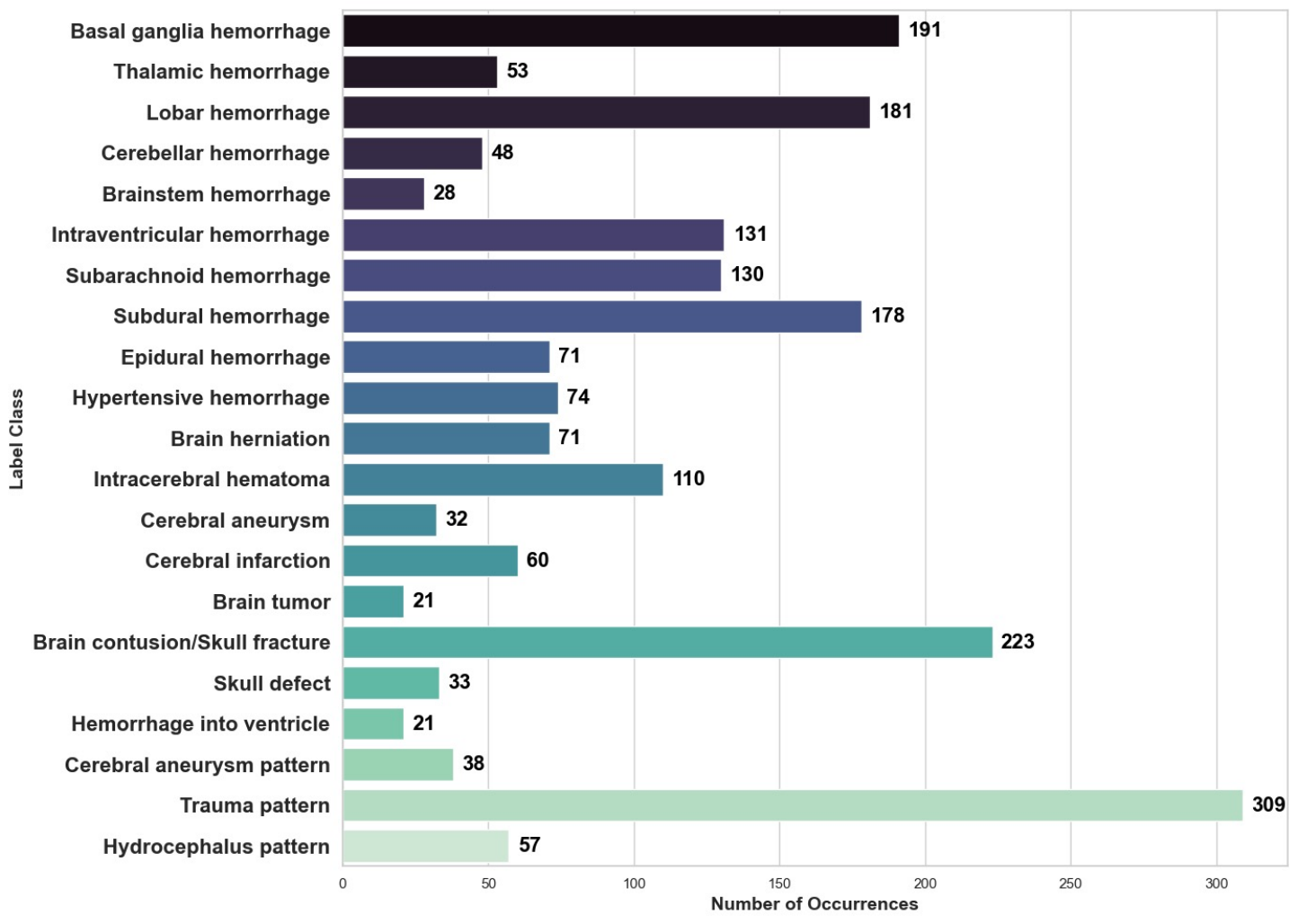}
\caption{Distribution of the brain pathological labels.}
\label{fig2}
\end{figure}

\noindent\textbf{Dataset and Preprocessing.}
We curated a real-world dataset of 852 non-contrast head CT (NCCT) studies, acquired from Siemens and GE scanners. Exclusion criteria comprised patients under 18 years of age and scans lacking axial reconstructions. A strict patient-level split allocated 682 cases for training and 170 for testing. Scans were axially aligned, resampled to $1{\times}1{\times}5$ mm, and resized to $224{\times}224{\times}32$. Hounsfield Units (HU) were clipped to $(0, 80)$ and normalized to $[0, 1]$. Training augmentations included random flipping, planar rotation ($\le 45^\circ$), and Gaussian noise ($\sigma=0.01$). Additionally, Fig.~\ref{fig2} illustrates the prevalence distribution of the LLM-extracted pathological labels.

\noindent\textbf{Implementation details.}
The proposed framework was built upon the pre-trained BiomedCLIP~\cite{biomedclip} architecture. LoRA adapters apply rank r = 16 with dropout probability 0.1 to the vision attention layers. Training were executed for 100 epochs on 2 NVIDIA GeForce RTX 4090 GPUs with a total batch size of 4. We use AdamW~\cite{adamw} as the optimizer, with a learning rate of 1e-05, a beta1 coefficient of 0.9, and a weight decay factor of 0.2.

\noindent\textbf{Comprehensive diagnosis of brain lesions.}
\begin{table*}[t]
\centering
\caption{Multi-label classification performance on the clinical dataset. Best and second-best results are highlighted in \textbf{bold} and \underline{underlined}, respectively. $\dagger$ denotes the variant optimized with standard BCE loss instead of ASL~\cite{asl}.}
\label{tab:multilabel_results}
\footnotesize
\renewcommand{\arraystretch}{0.9}
\setlength{\aboverulesep}{1.0pt}
\setlength{\belowrulesep}{1.5pt}
\setlength{\tabcolsep}{8pt}
\begin{tabular}{lcccccc}
\toprule
\multirow{2}{*}{\textbf{Method}}  
& \textbf{Micro} 
& \textbf{Micro} 
& \textbf{Micro} 
& \textbf{Macro}   
& \textbf{Hamming} \\
 & \textbf{SEN}
 & \textbf{SPE}  
 & \textbf{AUC} 
 & \textbf{AUC} 
 & \textbf{Loss} \\
\midrule
\rowcolor{gray!15} \multicolumn{6}{l}{\textit{3D Volumetric Models}} \\
ViT-B~\cite{vit}                    
& {0.567} & 0.780 & 0.759 & 0.534 & 0.240 \\
Swin-B~\cite{hatamizadeh2021swin}      
& \underline{0.612} & 0.802 & 0.778 & 0.535 & 0.131 \\
\midrule
\rowcolor{gray!15} \multicolumn{6}{l}{\textit{2D MIL Models (Trained from Scratch)}} \\
Mean Pooling        
& 0.266 & 0.977 & 0.775 & 0.493 & 0.094 \\
ABMIL~\cite{abmil}                
& 0.284 & 0.981 & 0.797 & 0.563 & 0.094 \\
TransMIL~\cite{shao2021transmil}            
& 0.328 & 0.956 & 0.819 & 0.749 & 0.103 \\
\midrule
\rowcolor{gray!15} \multicolumn{6}{l}{\textit{2D MIL Models (BiomedCLIP~\cite{biomedclip} Pre-trained)}} \\
Mean Pooling 
& 0.104 & \textbf{0.996} & 0.840 & 0.661 & 0.087 \\
ABMIL~\cite{abmil}        
& 0.400 & 0.965 & 0.860 & 0.750 & 0.088 \\
TransMIL~\cite{shao2021transmil}     
& 0.155 & \underline{0.993} & 0.857 & 0.698 & 0.086 \\
\midrule
\textbf{Brain-Adapter}$^{\dagger}$
& 0.285 & 0.982 & \underline{0.876} & \underline{0.773} & \underline{0.084} \\
\textbf{Brain-Adapter}   
& \textbf{0.740} & 0.865 & \textbf{0.887} & \textbf{0.778} & \textbf{0.079} \\
\bottomrule
\end{tabular}
\end{table*}
Table~\ref{tab:multilabel_results} compares Brain-Adapter against a 3D network (ViT-B~\cite{vit}, Swin-B~\cite{hatamizadeh2021swin}) and 2D MIL aggregators (Mean Pooling, ABMIL~\cite{abmil}, TransMIL~\cite{shao2021transmil}) trained from scratch or initialized with BiomedCLIP~\cite{biomedclip}. Native 3D transformers like ViT-B and Swin-B struggle severely, hampered by data scarcity and the inherent difficulty of capturing fine-grained details from thick-slice CTs without explicit medical priors. Conversely, BiomedCLIP pre-training substantially boosts all MIL baselines (e.g., ABMIL Macro AUC boosts from 0.563 to 0.750), validating the efficacy of foundation models for slice feature extraction. Ultimately, Brain-Adapter achieves state-of-the-art performance (Micro/Macro AUC: 0.887/0.778, Hamming Loss: 0.079). Furthermore, compared to the BCE variant ($\dagger$), employing Asymmetric Loss (ASL) prevents overwhelming true negatives from dominating gradients, drastically improving Micro Sensitivity from 0.285 to 0.740 in this highly imbalanced dataset.

\begin{figure}[t]
\centering
\includegraphics[width=\textwidth]{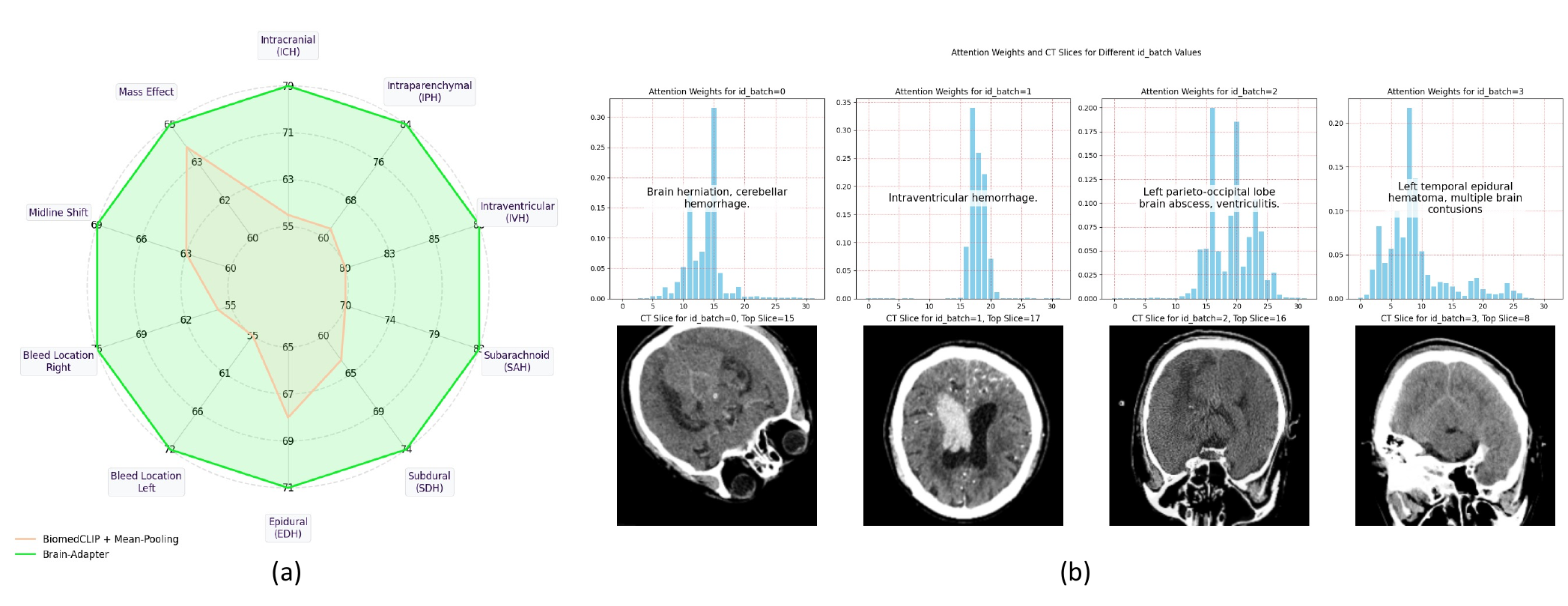}
\caption{Evaluation of zero-shot transferability and interpretability. \textbf{(a)} Zero-shot classification performance evaluated on the external CQ500~\cite{cq500} dataset. \textbf{(b)} Explainability analysis on our clinical dataset, visualizing the text-guided slice aggregation weights derived from TCA.}
\label{fig3}
\end{figure}

\noindent\textbf{Zero-shot abnormality detection.}
To evaluate cross-domain transferability, we explicitly leveraged the semantic grounding capability of TCA branch for a zero-shot detection task on the external CQ500~\cite{cq500} dataset (491 CT scans, ten abnormalities). By utilizing disease prompts as textual queries, Brain-Adapter consistently surpasses the BiomedCLIP Mean-Pooling baseline, as illustrated in Fig.~\ref{fig3} (a). This superiority indicates that our alignment strategy effectively preserves and enhances the zero-shot semantic reasoning of the 2D foundation model, adapting it to unseen 3D volumetric distributions without domain-specific retraining.

\noindent\textbf{Ablation study.}
\begin{table*}[!t]
\centering
\caption{Ablation study of our components: Open-set Alignment ($\mathcal{L}_{align}$), Consistency Constraint ($\mathcal{L}_{cons}$), and Uncertainty-Aware Refinement (UAR).}
\label{tab:ablation}
\footnotesize
\renewcommand{\arraystretch}{0.9}
\setlength{\aboverulesep}{1.0pt}
\setlength{\belowrulesep}{1.5pt}
\setlength{\tabcolsep}{4pt}
\begin{tabular}{ccc ccc}
\toprule
\multicolumn{3}{c}{\textbf{Components}} & \multicolumn{3}{c}{\textbf{Metrics}} \\
\cmidrule(r){1-3} \cmidrule(l){4-6}
$\mathcal{L}_{align}$ & $\mathcal{L}_{cons}$ & UAR & \textbf{Hamming Loss} ($\downarrow$) & \textbf{Micro AUC} ($\uparrow$) & \textbf{Macro AUC} ($\uparrow$)  \\
\midrule
\multicolumn{3}{c}{\textit{BiomedCLIP+ABMIL}} & 0.088 & 0.860 & 0.760 \\
\midrule
\checkmark & & & 0.089 & 0.873 & 0.772 \\
\checkmark & \checkmark & & 0.101 & 0.878 & 0.774 \\
\checkmark & & \checkmark & 0.083 & 0.879 & 0.776 \\
\midrule
\checkmark & \checkmark & \checkmark 
& \textbf{0.079} & \textbf{0.887} & \textbf{0.778} \\
\bottomrule
\end{tabular}
\end{table*}
Table~\ref{tab:ablation} details the ablation study validating the contribution of each proposed module. The sequential integration of components consistently drives performance upward. The Open-set Alignment loss ($\mathcal{L}_{align}$) forces the network to capture fine-grained textual semantics, while the consistency constraint ($\mathcal{L}_{cons}$) synchronizes the visual and text-guided representations. Finally, UAR mitigates ambiguous predictions during inference. The synergistic combination of all components yields the optimal configuration (Hamming Loss: 0.079), confirming the efficacy of the dual-stream architecture.


\noindent\textbf{Explainability analysis.}
Fig.~\ref{fig3}(b) visualizes the TCA slice-aggregation weights, demonstrating the inherent interpretability of our framework. The attention distributions reveal two vital clinical alignments. First, the model naturally prioritizes informative intermediate slices, correctly discarding diagnostically sparse extreme slices. Second, pathology-specific queries (e.g., \textit{"intraventricular hemorrhage"} of the second sample) precisely highlight the exact slices manifesting the lesion. This transparent semantic-visual grounding confirms that Brain-Adapter effectively captures deep clinical concepts, serving as a reliable diagnostic aid.

\section{Conclusion}
In this paper, we introduce Brain-Adapter, a novel dual-stream Vision-Language MIL framework that adapts 2D VLMs for the comprehensive 3D diagnosis of acute intracranial pathologies. To effectively leverage raw diagnostic reports, we construct an Open-set Alignment branch and a Logic-set Supervision branch. Furthermore, we propose an Uncertainty-Aware Refinement strategy to dynamically rectify visual predictions at test time using prompt-based semantic scores. Comprehensive experiments on a real-world clinical dataset establish a new state-of-the-art in multi-label head CT classification. Additionally, zero-shot evaluations on the external CQ500 dataset validate its cross-domain transferability. By bridging the dimensional gap between 2D foundation models and 3D clinical volumes, Brain-Adapter provides an accurate, interpretable, and scalable virtual assistant for critical care radiology.

\section*{Acknowledgments}
This research is supported by the National Natural Science Foundation of China (Grant No. 62471288).

%
%
\bibliographystyle{splncs04}
\bibliography{main}
\end{document}